\documentclass{article}
\usepackage[preprint]{neurips_2024}

\usepackage{natbib}
\setcitestyle{numbers,square}
\usepackage[utf8]{inputenc} 
\usepackage[T1]{fontenc}    
\usepackage{hyperref}       
\usepackage{url}            
\usepackage{booktabs}       
\usepackage{amsfonts}       
\usepackage{nicefrac}       
\usepackage{microtype}      
\usepackage{xcolor}         

\usepackage{algorithm}
\usepackage{lineno}
\usepackage{array}
\usepackage[caption=false,font=small,labelfont=sf,textfont=sf]{subfig}

\usepackage{textcomp}
\usepackage{stfloats}
\usepackage{verbatim}
\usepackage{graphicx}
\usepackage{etoolbox}
\usepackage{tabu}
\usepackage{ulem}
\usepackage{mathrsfs}

\usepackage{listings}
\usepackage{multicol}

\usepackage{pythonhighlight}

\usepackage{color}
\usepackage{epsfig}

\usepackage{subfiles}

\usepackage{adjustbox}

\usepackage{colortbl}
\usepackage{float}
\usepackage{hhline}
\usepackage{multirow}

\usepackage{bm}
\usepackage{mathtools}
\usepackage{changepage}
\usepackage{extramarks}
\usepackage{fancyhdr}
\usepackage{lastpage}
\usepackage{soul}
\usepackage{algpseudocode}
\usepackage{enumerate}
\usepackage{pbox}
\usepackage{footnote}
\usepackage{tablefootnote}
\usepackage{dsfont}
\usepackage{nccmath}

\usepackage[utf8]{inputenc} 
\usepackage[T1]{fontenc}    
\usepackage{booktabs}       
\usepackage{amsfonts}       
\usepackage{nicefrac}       
\usepackage{microtype}      
\usepackage{xcolor}         

\usepackage{algorithm}
\usepackage{lineno}
\usepackage{array}
\usepackage[caption=false,font=small,labelfont=sf,textfont=sf]{subfig}

\usepackage{textcomp}
\usepackage{verbatim}
\usepackage{etoolbox}

\usepackage{mathrsfs}

\usepackage{multicol}

\usepackage{pythonhighlight}

\usepackage{color}
\usepackage{epsfig}
\usepackage{amssymb}

\usepackage{subfiles}

\usepackage{adjustbox}

\usepackage{colortbl}
\usepackage{float}
\usepackage{hhline}
\usepackage{multirow}

\usepackage{bm}
\usepackage{mathtools}
\usepackage{changepage}
\usepackage{extramarks}
\usepackage{fancyhdr}
\usepackage{url}
\usepackage{soul}
\usepackage{algpseudocode}
\usepackage{enumerate}
\usepackage{pbox}
\usepackage{footnote}
\usepackage{tablefootnote}
\usepackage{dsfont}
\usepackage{nccmath}

\usepackage{stfloats}
\usepackage{tabu}
\usepackage{stmaryrd}
\usepackage{ulem}
\usepackage{listings}
\usepackage{subfiles}

\definecolor{mydarkgreen}{rgb}{0.02,0.6,0.02}
\definecolor{score_red}{RGB}{255, 100, 97}
\definecolor{score_green}{RGB}{0, 153, 50}

\definecolor{pink1}{rgb}{1, 0.3, 0.7}

\definecolor{babyblue}{rgb}{0.54, 0.81, 0.94}

\newcommand*{\belowrulesepcolor}[1]{%
  \noalign{%
    \kern-\belowrulesep
    \begingroup
      \color{#1}%
      \hrule height\belowrulesep
    \endgroup
  }%
}
\newcommand*{\aboverulesepcolor}[1]{%
  \noalign{%
    \begingroup
      \color{#1}%
      \hrule height\aboverulesep
    \endgroup
    \kern-\aboverulesep
  }%
}

\title{Is a Pure Transformer Effective for Separated and Online Multi-Object Tracking?}

\author{
  Chongwei Liu$^{1}$ \quad Haojie Li$^{1, 2}$ \quad Zhihui Wang$^{1}$ \quad Rui Xu$^{1*}$\\
  $^{1}$Dalian University of Technology \\
  $^{2}$Shandong University of Science and Technology \\
  \texttt{lcwdllg@mail.dlut.edu.cn \quad hjli@sdust.edu.cn} \\
  \texttt{zhwang@dlut.edu.cn \quad xurui@dlut.edu.cn} \\
}

\begin{document}

\maketitle

\begin{abstract}
Recent advances in Multi-Object Tracking (MOT) have demonstrated significant success in short-term association within the separated tracking-by-detection online paradigm. However, long-term tracking remains challenging. While graph-based approaches address this by modeling trajectories as global graphs, these methods are unsuitable for real-time applications due to their non-online nature.
In this paper, we review the concept of trajectory graphs and propose a novel perspective by representing them as directed acyclic graphs. This representation can be described using frame-ordered object sequences and binary adjacency matrices. We observe that this structure naturally aligns with Transformer attention mechanisms, enabling us to model the association problem using a classic Transformer architecture. Based on this insight, we introduce a concise Pure Transformer (PuTR) to validate the effectiveness of Transformer in unifying short- and long-term tracking for separated online MOT.
Extensive experiments on four diverse datasets (SportsMOT, DanceTrack, MOT17, and MOT20) demonstrate that PuTR effectively establishes a solid baseline compared to existing foundational online methods while exhibiting superior domain adaptation capabilities. Furthermore, the separated nature enables efficient training and inference, making it suitable for practical applications. Implementation code and trained models are available at \url{https://github.com/chongweiliu/PuTR}.
\end{abstract}
\section{Introduction}

Multi-Object Tracking (MOT) is a fundamental computer vision task essential for modern perception systems, ranging from autonomous driving~\cite{ess2010object} to video surveillance~\cite{elhoseny2020multi} and behavior analysis~\cite{hu2004survey}. The field has witnessed diverse methodological approaches, including motion estimation~\cite{sort, bytetrack} and graph-based techniques~\cite{jiang2007linear, mpntrack}.

Among these methodologies, the separated tracking-by-detection (TbD) online paradigm has become dominant due to its effectiveness and simplicity. This approach separates MOT into two distinct tasks: object detection and cross-frame association. Using ready-made detectors like YOLOX~\cite{yolox}, it treats frame-by-frame detection results as immutable and focuses on solving the association problem. Methods such as SORT~\cite{sort} and ByteTrack~\cite{bytetrack} demonstrate this approach's effectiveness, using Kalman filtering~\cite{kalman} for motion prediction and the Hungarian algorithm~\cite{hungarian} for Intersection-over-Union (IoU) based association.
While these handcrafted heuristic approaches excel in short-term tracking scenarios without data-driven learning, they inherently struggle with long-term tracking when relying solely on motion information.

The long-term tracking — maintaining object identity through extended occlusions or absences — remains a significant challenge for above heuristic approaches. These methods lack a unified framework for modeling both short- and long-term associations, often resorting to complex handcrafted rules or additional components~\cite{ocsort, hybridsort} for long-term tracking. In response, graph-based methods~\cite{mpntrack, sushi} have emerged, modeling trajectories as global graphs and employing Graph Neural Networks (GNNs)~\cite{gnn} for association. While these approaches effectively handle both short- and long-term scenarios through global context modeling, they sacrifice online processing capability, limiting their real-time applicability.

\begin{figure*}[t]
    \centering
    \includegraphics[width=0.95\textwidth]{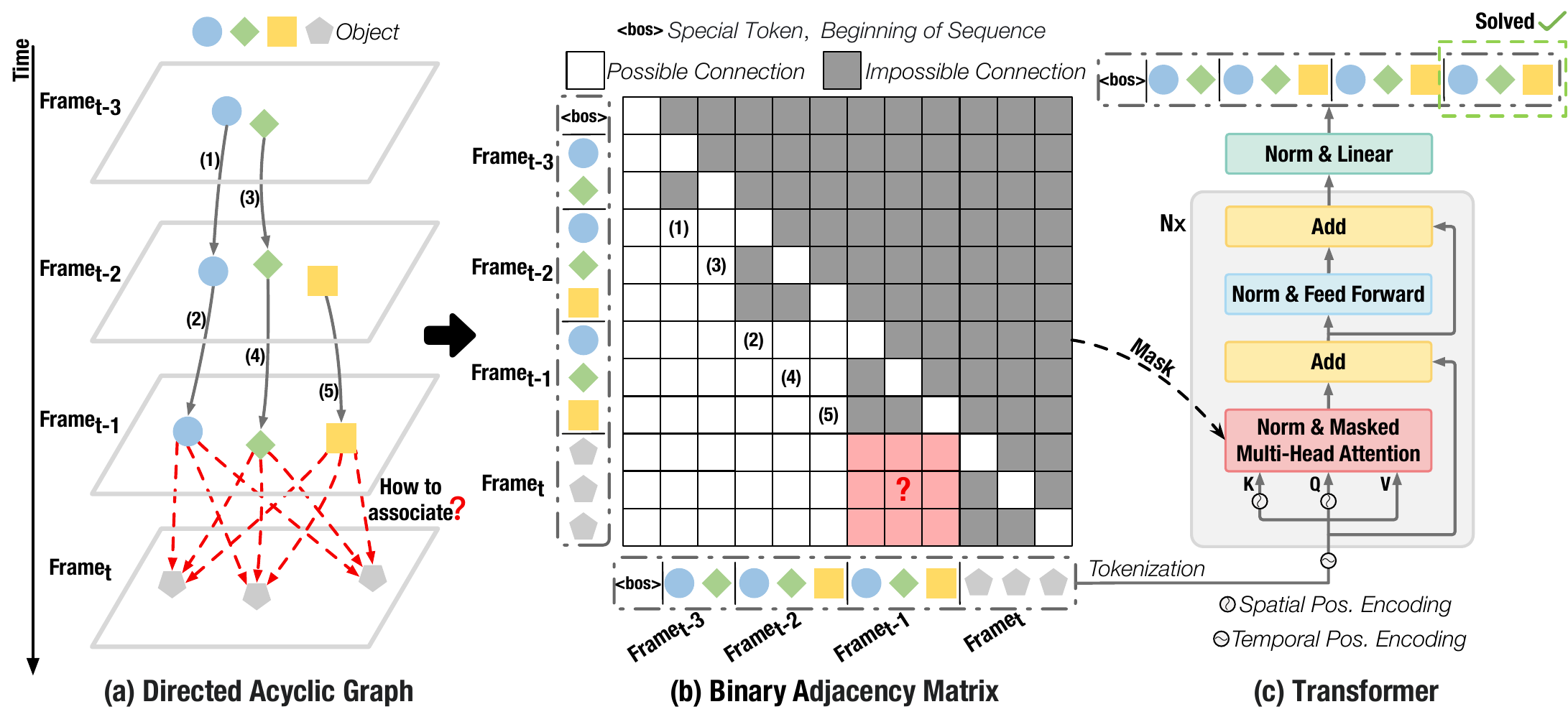}
    \caption{Illustration of our chain of thought. Trajectories are inherently a directed acyclic graph in the temporal order (a). We can thus transform it equivalently into a binary adjacency matrix (b), exactly aligning with the Transformer's attention mask. Consequently, arranging objects by frame forms a natural input sequence for the Transformer (c), enabling it to model the association problem.}
    \label{fig:intro}
    \vspace{-0.6cm}
\end{figure*}

This leads us to a fundamental question: How can we elegantly unify short- and long-term tracking challenges while maintaining a separated and online manner? Our key insight, illustrated in Fig.\ref{fig:intro}, stems from recognizing that trajectory graphs are inherently directed acyclic graphs (DAGs), i.e., each trajectory propagates independently in time, with nodes constrained to at most one in-degree and one out-degree. This structure enables representation as a frame-ordered object sequence with a binary adjacency matrix (Fig.\ref{fig:intro}b). Notably, this matrix's structure — comprising white blocks for possible connections and gray blocks for impossible ones — mirrors the attention mask in Transformer architectures~\cite{transformer} (Fig.\ref{fig:intro}c).
This structural alignment motivates our intuition: trajectory graphs can be naturally modeled via a pure Transformer in an online and separated paradigm, complementing traditional non-online graph-based methods.


Previous works like TrackFormer~\cite{trackformer} and MOTR~\cite{motr} have already incorporated Transformers into MOT by adapting the Transformer-based detector DETR~\cite{detr} for end-to-end tracking. The key difference between these works and our idea is the track propagation mechanism. They modify Deformable-DETR~\cite{deformabledetr} to propagate tracklets across frames via query tokens to perform detection and association simultaneously. This propagation trait stems from earlier works like Tracktor~\cite{tracktor} and CenterTrack~\cite{centertrack}, which utilized CNN-based detectors~\cite{fasterrcnn, centernet} at that time. However, their track propagation mechanism more closely resembles RNN~\cite{rnn} behavior rather than a true Transformer, and the Transformer structure is only used for detection. \textit{In these cases, the true power of the self-attention mechanism to model context dependencies is not exploited for association tasks. Our straightforward idea is to leverage the potential of Transformers for the association problem, thereby opening up a new and promising research direction in the field of MOT.}

In this paper, we present \underline{\textbf{Pu}}re \underline{\textbf{Tr}}ansformer (PuTR), a concise architecture that validates our intuitive idea. Following Occam's Razor, we utilize established practices for adapting the Transformer to MOT. As Fig.\ref{fig:intro}c shows, we modify the attention mask and incorporate temporal and spatial positional encodings to adapt it to the unique characteristics of the object sequence compared to a normal text sequence. Specifically, unlike a sentence, the object sequence is ordered in the outer frame temporal dimension but unordered in the inner image spatial dimension. Therefore, we modify the attention mask to ensure permutation invariance within the same frame and introduce temporal and spatial positional encodings to encode the frame order and the object coordinate information into the object sequence. We conduct experiments on four MOT datasets: SportsMOT~\cite{sportsmot}, DanceTrack~\cite{dancetrack}, MOT17~\cite{mot17}, and MOT20~\cite{mot20}, to demonstrate the effectiveness of our approach. The results show that PuTR achieves a solid baseline compared to existing foundational online methods. Notably, PuTR exhibits superior domain adaptation ability, with a maximum cross-dataset gap of only $1.7\%$ in HOTA without fine-tuning. Furthermore, PuTR is efficient in both training and inference. It can be trained from scratch on a single GPU in one hour without additional data or complex training processes, and achieves up to 77 FPS in inference when given detection results.

In summary, our main contributions are:
\begin{itemize}
    \item We review the trajectory graph and discover that it naturally aligns with the Transformer architecture, providing a theoretical foundation for applying Transformers to solve the association problem in multi-object tracking.
    \item We propose a concise Pure Transformer architecture (PuTR) to validate our intuitive idea. By inputting the object sequence in temporal order, our approach maintains online processing capability while naturally unifying short- and long-term object associations.
    \item We conduct experiments on four datasets to demonstrate the effectiveness of our approach in terms of performance, domain adaptation, and efficiency. The results show that PuTR achieves a solid baseline compared to existing foundational online methods.
\end{itemize}

\section{Related Work}
The MOT field has witnessed diverse approaches to address the challenging association problem. We provide a concise overview of the most representative methodologies.

\subsection{Heuristic Methods}
Heuristic methods constitute the dominant paradigm in MOT. Following the separated paradigm, these approaches first employ pre-trained detectors~\cite{fasterrcnn, yolox, centernet} for frame-wise object detection, followed by online cross-frame association. Given the fixed nature of detection results, research efforts primarily concentrate on the association problem, with particular emphasis on motion modeling and spatial relationships.

SORT~\cite{sort} established the foundational framework by integrating Kalman filtering\cite{kalman} for motion prediction with the Hungarian algorithm\cite{hungarian} for association, while introducing a runtime manager system for tracklets (\textit{Track}, \textit{Lost}, and \textit{New}). Subsequent research has largely built upon SORT's architecture, enhancing association robustness through various strategies: integrating appearance features via dedicated ReID models~\cite{deepsort, strongsort, hybridsort, you2024multi}, developing advanced motion models~\cite{bytetrack, ocsort, huang2024deconfusetrack}, and incorporating camera motion compensation~\cite{botsort, deepocsort, yi2024ucmctrack}.

Despite their effectiveness in short-term tracking scenarios, these approaches face inherent limitations in unifying short-term and long-term tracking within a single theoretical framework. Maintaining tracklet completeness in long-term scenarios often requires additional complex handcrafted rules, highlighting the fundamental constraints of this methodology.

\subsection{Graph-based Methods}
Graph-based approaches have emerged as a promising alternative to address the inherent limitations of heuristic methods. Within the separated paradigm, these approaches formulate MOT as a graph optimization problem, where detected objects serve as nodes and trajectory hypotheses as edges. Unlike traditional frame-by-frame trackers, graph-based methods pursue global solutions for data association across multiple frames or entire sequences, achieving robust performance in semi-online or offline settings.

Early research in this domain explored various optimization frameworks, including fixed cost minimization~\cite{jiang2007linear, 5995604}, multi-cuts~\cite{tang2017multiple}, and minimum cliques~\cite{roshan2012gmcp}. Recent advances in GNNs have introduced more sophisticated approaches~\cite{mpntrack, sushi}. MPNTrack~\cite{mpntrack} leverages message passing networks for detection association through edge classification, while SUSHI~\cite{sushi} implements recursive hierarchical levels to optimize graph space overhead and process extended video sequences. While these methods successfully unify short- and long-term tracking under a global perspective, their offline nature presents limitations for real-time applications.

Our key contribution stems from recognizing the inherent structure of trajectory graphs as DAGs. This insight enables a novel approach to modeling association problems using Transformers in a separated and online manner, effectively balancing comprehensive context modeling with real-time processing capabilities.

\subsection{Detector-based Methods}
Given that object detection is prerequisite for the TbD paradigm, efforts have emerged to integrate detection and tracking into unified, simultaneously trainable frameworks. Joint-Detection-and-Embedding (JDE)~\cite{jde} and FairMOT~\cite{fairmot} represent transitional approaches, modifying CNN-based detectors to generate both object boxes and appearance features for heuristic association~\cite{guo2023feature, qdtrack}.

In the CNN era, unified frameworks like Tracktor~\cite{tracktor} and CenterTrack~\cite{centertrack} modified detectors to predict current tracklets from previous results~\cite{wang2023jdan, sun2019deep}. The Transformer era, dominated by the DETR family~\cite{detr, deformabledetr, dabddetr}, saw TrackFormer~\cite{trackformer} and MOTR~\cite{motr} extend Deformable-DETR's~\cite{deformabledetr} object queries to MOT. Recent advances focus on historical object embedding storage~\cite{memot, memotr} and resolving conflicts between new and tracked object queries~\cite{comot, motrv3}. Despite architectural evolution from CNN to Transformer, the core mechanism remains RNN-style tracklet state propagation. PuTR explores an alternative approach, focusing on association rather than detection.

MOTIP~\cite{motip} similarly recognizes Transformers' association potential, formulating tracking as end-to-end in-context ID prediction. It employs a learnable ID dictionary and Transformer decoder post-Deformable-DETR for tracking ID prediction, using historical trajectories as $K$ and $V$, and current detections as $Q$ in attention mechanisms, where $K$, $Q$, and $V$ denote the \textit{Key}, \textit{Query}, and \textit{Value} matrices, respectively. While conceptually similar, PuTR and MOTIP differ significantly in implementation. PuTR's separated, decoder-only architecture enables efficient single-GPU training without pre-training or complex stabilization. Conversely, MOTIP's end-to-end approach requires multi-GPU training and empirical ID dictionary tuning, while PuTR chooses learning assignment matrices directly \cite{xu2020train} for relative affinity association to avoid such constraints, achieving superior domain adaptation.

We acknowledge that these methodologies are complementary rather than competitive. The runtime manager appears in DETR-based methods, while our approach draws from GNN-based trajectory graphs. MOTIP bridges MOTR and PuTR, with each method occupying distinct niches. End-to-end frameworks excel with sufficient video-level data (e.g., DanceTrack), while our separated approach provides a baseline for scarce-data scenarios (e.g., MOT17). We advocate for methodological coexistence, offering comprehensive solutions to the MOT community.

\section{Method}
This section introduces the implementation of PuTR, which leverages the Transformer's self-attention mechanism to model the association problem in separated and online MOT. Section~\ref{sec:preliminary} presents the fundamental concepts and problem formulation of MOT and introduces the input object sequence and tokenization process. Section~\ref{sec:architecture} provides details of the PuTR architecture, including modifications on causal masking and positional encoding. Sections~\ref{sec:training} and~\ref{sec:inference} describe the training and inference processes, respectively.

\subsection{Preliminary}
\label{sec:preliminary}
We first introduce the fundamental concepts in PuTR. Our approach follows the TbD paradigm. Given a video, we assume that object detections are computed for every frame, and our task is to perform data association by linking object detections into online trajectories.

\subsubsection{Problem Formulation}
We denote the set of objects in the $t$-th frame as $\mathcal{O}_{t} = \{o_\textit{tid}^{t}\}$, where $o_\textit{tid}^{t}$ represents the detected object with the tracking identity $\textit{tid}$. Each $o_\textit{tid}^{t}$ is represented by its bounding box (bbox), confidence score, and category label. The $\textit{tid}$ is assigned to each object belonging to the same trajectory, with $\textit{tid}=-1$ indicating that the object does not belong to any trajectory. Within an $\mathcal{O}_{t}$ set, each $\textit{tid}$ is unique, except for $-1$. The goal of association is to obtain the set of trajectories $\mathcal{T}^*$ that group detections corresponding to the same identity. Each trajectory $\mathcal{T}_k \in \mathcal{T}^*$ is given by its set of detections $\mathcal{T}_k = \{o_{k}^{t} \ | \ t_{s,k} \leq t \leq t_{e,k}\}$, where $t_{s,k}$ and $t_{e,k}$ are the start and end frames of the trajectory $\mathcal{T}_k$, respectively.

\subsubsection{Input Object Sequence}
Following Transformer manner, the PuTR input is also a sequence, but consisting of detections in recent $T+1$ consecutive frames. We denote the input object sequence as $\mathcal{S} = ({\fontfamily{phv}\selectfont\textsf{<bos>}}, \mathcal{O}_{t-T}, ..., \mathcal{O}_t)$. Similar to large language models, $\mathcal{S}$ starts with a special token {\fontfamily{phv}\selectfont\textsf{<bos>}} to indicate the beginning of the sequence, which is an all-zero token. The input sequence is then fed into PuTR after tokenization.

\subsubsection{Tokenization}
To feed into PuTR, we first convert each object $o_\textit{tid}^{t}$ into a token representation. We denote the token representation of $o_\textit{tid}^{t}$ as $x_\textit{tid}^{t}$ and the corresponding token set as $\mathcal{X}_{t} = \{x_\textit{tid}^{t}\}$. It is feasible to map objects into high-dimensional vectors by employing an independent Re-ID model~\cite{strongsort} or DETR object embeddings~\cite{motr}. However, in this work, to maintain simplicity and efficiency, we adopt a straightforward approach by directly encoding the object image patches from the raw image into token representations. Specifically, for each object image patch $p_\textit{tid}^{t} \in \mathbb{R}^{h \times w \times 3}$, generated by cropping the corresponding object detection bbox\footnote{$h$ and $w$ are the height and width of the bbox.} from the $t$-th frame (raw image), we divide it into a $64 \times 64$ grid and sample the center point of each grid via bilinear interpolation. We then flatten these points to obtain a 1-d representation $r_\textit{tid}^{t} \in \mathbb{R}^{1 \times (64 \cdot 64 \cdot 3)}$, and pass it through a linear layer to obtain the tokenized representation $x_\textit{tid}^{t}$ with the same dimension as the PuTR model dimension $d_\textit{model} = 512$. The tokenized sequence thus becomes $\mathcal{S}_\textit{in} = ({\fontfamily{phv}\selectfont\textsf{<bos>}}, \mathcal{X}_{t-T}, ..., \mathcal{X}_t) \in \mathbb{R}^{n_{\mathcal{S}} \times d_\textit{model}}$, where $n_{\mathcal{S}}$ is the token number of the sequence.

\subsection{Architecture}
\label{sec:architecture}
\subsubsection{Overview}
As shown in Fig.\ref{fig:intro}c, PuTR follows the standard Transformer architecture, composed of a stack of $N=6$ identical layers. Each layer has two sub-layers: a masked multi-head self-attention mechanism with $8$ heads, and a feed-forward network. A residual connection is employed around each sub-layer. To improve training stability, we perform normalization before each sub-layer input following Llama2~\cite{llama2}. The $\mathcal{S}_\textit{in}$ is fed into PuTR after adding temporal positional encodings, and spatial positional encodings are added before being linearly projected into $K$ and $V$ in each attention mechanism. After the $N$ layers, the output object embeddings are linearly projected into high-dimensional representations, denoted as $\mathcal{S}_\textit{out} = ({\fontfamily{phv}\selectfont\textsf{<bos>}}, \mathcal{Z}_{t-T}, ..., \mathcal{Z}_t) \in \mathbb{R}^{n_{\mathcal{S}} \times d_\textit{model}}$, which is then used to calculate the relative affinity matrix between trajectories and current detections to solve the association problem.

\subsubsection{Modifications}
Unlike sentences, object sequences are ordered temporally across frames but unordered spatially within each frame. To account for this, we modify the attention mask and introduce temporal and spatial positional encodings.

Since objects in $\mathcal{O}_{t}$ are unordered and cannot be interlinked, we modify the conventional causal mask (a lower triangular matrix) to ensure permutation invariance within the same frame. Specifically, based on the causal mask, we additionally set the positions to zero for tokens in the same frame, i.e., $\textit{mask}_{ij} = 0$ if $t_i == t_j \ \& \  i \neq j$, where $t_i$ and $t_j$ are the frame indices of the $i$-th and $j$-th tokens, respectively. We refer to this as the frame causal mask, which precisely aligns with the distribution of the white blocks in Fig.\ref{fig:intro}b, and thus is consistent with the trajectory graph structure.

For positional encoding, we employ two distinct approaches. Temporally, we utilize the absolute positional encoding from the original Transformer to encode the frame order. Spatially, we adopt the method from DAB-DETR~\cite{dabddetr} to encode coordinate information. As illustrated in Fig.\ref{fig:intro}c, tokens within $\mathcal{X}_{t}$ are augmented with identical temporal positional encodings before the first layer, denoting their frame order within the current clip (e.g., $\mathcal{X}_{t-T}$ at position 0, $\mathcal{X}_{t}$ at position $T$). Prior to linear projection to $K$ and $V$, each token is added with a spatial positional encoding derived from its own coordinates, thereby emphasizing spatial location information.

\subsection{Training}
\label{sec:training}
Unlike large language model pre-training, where each output token is required to predict the next word in a fixed vocabulary table via classification, in MOT, the same individual object generates a unique trajectory, making it impractical to establish a fixed individual object vocabulary. While MOTIP proposes using a shared tracking ID dictionary to predict the ID labels of current detected objects, determining the appropriate dictionary size requires empirical tuning for different datasets, and there is a risk of ID exhaustion when the number of tracked objects exceeds the dictionary size.

To ensure generalizability, PuTR circumvents the ID dictionary and instead leverages the relative affinity matrix to classify the current detected objects, thus accomplishing the association task. Specifically, we calculate the inner product between $\mathcal{Z}_{t-1}$ and $\mathcal{Z}_{t}$, i.e., $\mathcal{Z}_{t-1}\mathcal{Z}_{t}^\top$ $(t\geq 1)$, to obtain the affinity matrix, and apply the softmax along rows to get the probability distribution of the previous objects corresponding to the next objects. Through this, we can determine the association via the Hungarian algorithm, similar to heuristic methods, without the need for a fixed vocabulary table. The classification loss is therefore simply calculated as the cross-entropy loss between the predicted probability and the ground truth label. Without bells and whistles, the training loss of PuTR is straightforwardly calculated as:
\begin{equation}
\mathcal{L} =
\frac{\sum\limits^\textit{bs}\sum\limits^{T}\textit{CE}(\textit{softmax}_r(\mathcal{Z}_{t-1}\mathcal{Z}_{t}^\top), Y_t)}
{\textit{bs} \times T},
\label{Eq:Loss}
\end{equation}
where $\textit{CE}(\cdot)$ denotes the cross-entropy loss, $Y_t$ is the ground truth label in the $t$-th frame, and $\textit{bs}$ is the batch size. $\textit{CE}(\cdot)$ calculates the cross-entropy loss between the predicted probability and the ground truth label, averaging over the number of objects in $\mathcal{Z}_{t-1}$. For an identity $\textit{tid}$ in $\mathcal{Z}_{t}$ but not in $\mathcal{Z}_{t-1}$, we assign the nearest output object embedding with the same $\textit{tid}$ prior to $\mathcal{Z}_{t-1}$. This strategy adds long distance training samples to overcome object intervals. To prevent anomalies in loss calculation, no row corresponds to an object in $\mathcal{Z}_{t}$ that initiates its trajectory, and we also remove rows with $\textit{tid}$ in $\mathcal{Z}_{t-1}$ but not in $\mathcal{Z}_{t}$.

\subsection{Inference}
\label{sec:inference}
Different methodologies are not mutually exclusive. Although DETR-based methods are considered end-to-end, some of them still require a runtime manager to schedule tracklet states~\cite{motrv2} and occasionally resolve assignment conflicts between tracklets and detections via heuristic rules or the Hungarian algorithm during inference~\cite{motip}. From this perspective, they can also be near "Heuristic Methods," with the distinction that the similarity matrix is generated by neural networks instead of handcrafted models. Therefore, it is natural that our proposed PuTR, as a separated framework, incorporates a runtime manager and the Hungarian algorithm to deal with similar issues during inference. Notably, we only match once per frame, emphasizing the Transformer's capability in solving the association problem, instead of using multiple matching strategies like ByteTrack.

\subsubsection{Pipeline}
In the first frame of a video sequence during inference, detected objects with a confidence score greater than the tracklet threshold $\tau_{\textit{new}}$ are recorded as newborn objects and assigned unique tracking identities. At each subsequent time step $t$, we first filter the detection results with a detection threshold $\tau_{\textit{det}}$. These remaining detections $\mathcal{O}_{t}$ are then concatenated with the object sequence $\mathcal{S}$ from the recent $T$ frames and fed into PuTR to generate an affinity matrix. We then employ the Hungarian algorithm to obtain assignment results and a runtime manager to update the tracklet states. The unmatched object with a detection confidence greater than $\tau_{\textit{new}}$ will be considered a newborn target and given a new identity. Please refer to Appendix~\ref{app:pipe} for more details.

\subsubsection{Affinity Matrix}
PuTR calculates distances between object features of different frames instead of identifying explicit individuals due to the absence of an ID dictionary. This approach, however, inevitably disregards spatial positioning even after adding spatial positional encodings, leading to erroneous matches between visually similar but spatially distant objects. Moreover, occlusions or illumination changes may cause objects from the previous frame ($t-1$) to vanish or lose discriminative information. To address these issues, we compute the similarity matrix $\mathbf{S}$ on $\mathcal{S}_\textit{out}$ rather than just the last two frames, and incorporate bboxes to compensate for the lack of spatial information. The matrix $\mathbf{S}$ is calculated as:
\begin{equation}
\mathbf{S} = \textit{softmax}_{r}([\mathcal{Z}_{t-T}, ..., \mathcal{Z}_{t-1}] \mathcal{Z}_t^\top) \cdot \mathbf{I}^{\alpha} + \mathbf{I}^{\beta},
\end{equation}
where $\mathbf{I}^{\alpha}$ and $\mathbf{I}^{\beta}$ are IoU matrices to compensate for the similarity in object scale and position, respectively. We compute $\mathbf{I}^{\alpha}$ between the bboxes of $[\mathcal{O}_{t-T}, ..., \mathcal{O}_{t-1}]$ and detections $\mathcal{O}_t$. To focus solely on object scale, we calculate IoU after setting the center coordinates of all bboxes to zero while preserving their width and height. $\mathbf{I}^{\beta}$ is the standard IoU between trajectory bboxes\footnote{The bbox/confidence score of a trajectory is from the object with the largest frame index in the same \textit{tid}, following heuristic methods.} of $[\mathcal{O}_{t-T}, ..., \mathcal{O}_{t-1}]$ and $\mathcal{O}_t$ to focus on the object position. Please refer to Appendix~\ref{app:similarityc} for detailed explanations. Each row of $\mathbf{S}$ represents the similarity between an object in previous frames and all objects in the current frame. We then take the maximum value under the same trajectory for each detection to obtain the trajectory-detection affinity matrix $\mathbf{A}$, where each row represents the similarity between a trajectory and all current detections. The maximum operation ensures the overall similarity is governed by the most distinctive pair, enhancing robustness. Inspired by heuristic methods, we use a weight matrix $\mathbf{W}$ to further prune invalid matches from bipartite matching. $\mathbf{W}$ has the same shape as $\mathbf{A}$ and is calculated as:
\begin{equation}
\mathbf{W}_{ij} = \mathbf{H}_{ij} \cdot \mathbf{T}_{i} \cdot \mathbf{D}_{j}
\end{equation}
where $\mathbf{H}$ is the Height Modulated IoU from Hybrid-SORT~\cite{hybridsort}, calculated using trajectory and detection bboxes; $\mathbf{T}_{i}$ and $\mathbf{D}_{j}$ are the confidence scores of the $i$-th trajectory and $j$-th detection, following FastTrack~\cite{fasttrack}. Finally, we employ the Hungarian algorithm on $\mathbf{A} \cdot \mathbf{W}$ to obtain the final assignment results.

\section{Experiments}
\subsection{Settings}
\label{sec:expersettings}

\subsubsection{Datasets}
For comprehensive analysis, we evaluate PuTR on four MOT benchmarks: MOT17, MOT20, DanceTrack, and the recently proposed SportsMOT. These datasets represent diverse tracking scenarios with distinct characteristics. MOT17 and MOT20 focus on pedestrian tracking with relatively regular motion patterns. MOT17 typically contains sparse pedestrians at small scales, while MOT20 features dense crowds (over 100 people per frame) at larger scales. DanceTrack presents challenging scenarios with performers executing complex movements on stage, characterized by frequent directional changes and mutual occlusions. SportsMOT encompasses various athletic scenarios (basketball, football, volleyball), featuring rapid movements and variable object speeds between adjacent frames.

The test sets of MOT17 and MOT20 contain limited sequences ($7$ and $4$, respectively), which has led to practices like per-sequence threshold tuning and post-interpolation in heuristic methods. Such practices may compromise the reliability of performance metrics. In contrast, SportsMOT and DanceTrack provide substantially larger test sets ($35$ and $150$, respectively), making such optimizations impractical and yielding more dependable evaluation results.

\subsubsection{Metrics} We employ three widely-used metrics: HOTA~\cite{hota}, IDF1~\cite{idf1}, and MOTA~\cite{mota}. MOTA measures object coverage, IDF1 focuses on identity preservation, and HOTA strikes a balance between these two aspects. Within HOTA, AssA assesses association performance, while DetA evaluates detection performance.
When methods are evaluated on the same detection results, MOTA values are nearly identical across methods. HOTA and IDF1 can better reflect the performance of the association component.

\subsubsection{Implementation Details}
To enhance the efficiency, we adopt several strategies when establishing PuTR with PyTorch.
We leverage the \texttt{grid\_sample} function\footnote{\fontsize{8pt}{12pt}\selectfont\texttt{torch.nn.functional.grid\_sample}} to simultaneously sample the grid center points of all objects within a frame. Moreover, the \texttt{scaled\_dot\_product\_attention} function\footnote{\fontsize{8pt}{12pt}\selectfont\texttt{torch.nn.functional.scaled\_dot\_product\_attention}} is employed to accelerate the calculation of the attention mechanism during both training and inference. Additionally, we implement a runtime manager following FastTrack~\cite{fasttrack}, maintaining active tracklets in a tensor array instead of a list of Python entities, thereby facilitating parallel processing and further improving efficiency.

Both training and inference of PuTR are conducted on an RTX 4090 GPU.
During training, consistent settings are maintained across all datasets for simplicity.
PuTR is optimized using the AdamW optimizer and a cosine learning rate schedule with a batch size of $4$ for $7$ epochs on the corresponding training set.
The length of training clips is progressively increased from $4$ to $8$, $16$, $32$, $64$, and $128$ at the $2$nd, $3$rd, $4$th, $5$th, and $6$th epoch, respectively, except on MOT20, where the clip length stays at $64$ after the $6$th epoch due to GPU memory limitations.
The initial learning rate is set to $0.0002$. We apply weight decay of $0.0005$, gradient clipping of $1.0$, and accumulate gradients for $32$ steps. Following the practices of previous works~\cite{motrv2, memotr}, training clips are sampled from the training sequences with a random interval ranging from $1$ to $10$, and augmented with random cropping, random resizing, HSV color jittering, and random horizontal flipping to enhance diversity.

During inference, we adopt publicly available YOLOX detections for a fair comparison. The detection results for DanceTrack are obtained from \url{https://github.com/DanceTrack/DanceTrack}, for MOT17 and MOT20 from \url{https://github.com/ifzhang/ByteTrack}, and for SportsMOT from \url{https://github.com/MCG-NJU/MixSort}. The consecutive frame $T$ is set to $30$ for all datasets.
Hyperparameters are consistent within the same dataset and not specifically tuned for each test sequence.
We maintain consistent hyperparameters within each test set \textit{without per-sequence threshold tuning and post-interpolation}.
In the inference stage, we set the confidence threshold $\tau_{\textit{det}}$ to $0.1$ across all four datasets. The newborn threshold $\tau_{\textit{new}}$ is set to $0.6$ for SportsMOT and DanceTrack, $0.7$ for MOT17, and $0.4$ for MOT20, generally following the practices of previous works~\cite{bytetrack, ocsort}.

\begin{table}[t]
    \small
    \centering
    \caption{Ablations on the affinity matrix. The Compensation, $\mathbf{H}$, $\mathbf{T}$, and $\mathbf{D}$ indicate the use of the similarity compensation, Height Modulated IoU, trajectory confidence score, and detection confidence score, respectively.}
    \resizebox{0.8\columnwidth}{!}{
    \begin{tabular}{
        cccc
        |c@{\hspace{1cm}}c@{\hspace{1cm}}c
        |c@{\hspace{1cm}}c@{\hspace{1cm}}c}
        \toprule
        \multicolumn{4}{c|}{Affinity Matrix}
        & \multicolumn{3}{c|}{SportsMOT val}
        & \multicolumn{3}{c}{DanceTrack val}  \\
        Compensation & $\mathbf{H}$ & $\mathbf{T}$ & $\mathbf{D}$
        & HOTA$\!\shortuparrow$ & MOTA$\!\shortuparrow$ & IDF1$\!\shortuparrow$
        & HOTA$\!\shortuparrow$ & MOTA$\!\shortuparrow$ & IDF1$\!\shortuparrow$ \\
        \midrule
        &&&& 76.8 & 95.7 & 78.7 & 52.3 & 88.7 & 52.8 \\
        $\checkmark$ &&&& 78.4 & 95.8 & 80.7 & 55.0 & 88.9 & 55.5 \\
        $\checkmark$ & $\checkmark$ &&& 78.7 & 95.8 & 81.0 & 55.4 & 89.1 & 56.0 \\
        $\checkmark$ & $\checkmark$ & $\checkmark$ && 78.7 & 95.8 & 81.1 & 56.0 & 89.2 & 57.4 \\
        $\checkmark$ & $\checkmark$ && $\checkmark$ & 78.9 & 95.8 & 81.2 & 57.3 & 89.3 & 58.7 \\
        $\checkmark$ & $\checkmark$ & $\checkmark$ & $\checkmark$
        & \textbf{78.9} & \textbf{95.8} & \textbf{81.2}
        & \textbf{57.6} & \textbf{89.4} & \textbf{59.4} \\
        \bottomrule
    \end{tabular}}
    \label{tab:am}
\end{table}

\begin{table}[t]
    \small
    \centering
    \caption{Ablations on the attention mask. Mask denotes the type of mask in the self-attention mechanism, and Shuffle indicates whether the object sequence of the current frame is shuffled during inference.}
    \resizebox{0.8\columnwidth}{!}{
    \begin{tabular}{c@{\hspace{1cm}}c
    |c@{\hspace{1cm}}c@{\hspace{1cm}}c
    |c@{\hspace{1cm}}c@{\hspace{1cm}}c}
        \toprule
        \multicolumn{2}{c|}{PuTR} & \multicolumn{3}{c|}{SportsMOT val} & \multicolumn{3}{c}{DanceTrack val} \\
        Mask & Shuffle
        & HOTA$\!\shortuparrow$ & MOTA$\!\shortuparrow$ & IDF1$\!\shortuparrow$
        & HOTA$\!\shortuparrow$ & MOTA$\!\shortuparrow$ & IDF1$\!\shortuparrow$ \\
        \midrule
        causal &
        & 77.9 & 95.6 & 80.3
        & 55.7 & 88.8 & 57.7 \\
        causal & $\checkmark$
        & 77.5 & 95.7 & 80.0
        & 55.9 & 88.8 & 57.9 \\
        \midrule
        frame causal &
        &\bf  78.9 & \bf 95.8 & \bf 81.2
        & \bf 57.6 & \bf 89.4 & \bf 59.4 \\
        frame causal & $\checkmark$
        & 78.9 & 95.8 & 81.2
        & 57.6 & 89.4 & 59.4 \\
        \bottomrule
    \end{tabular}}
    \label{tab:mask}
\end{table}

\begin{table}[t]
    \small
    \centering
    \caption{Ablations on the positional encoding. Temporal and Spatial denote the use of temporal positional encoding and spatial positional encoding, respectively.}
    \resizebox{0.8\columnwidth}{!}{
    \begin{tabular}{c@{\hspace{1cm}}c
    |c@{\hspace{1cm}}c@{\hspace{1cm}}c
    |c@{\hspace{1cm}}c@{\hspace{1cm}}c}
        \toprule
        \multicolumn{2}{c|}{Positional Encoding} & \multicolumn{3}{c|}{SportsMOT val} & \multicolumn{3}{c}{DanceTrack val}  \\
        Temporal & Spatial
        & HOTA$\!\shortuparrow$ & MOTA$\!\shortuparrow$ & IDF1$\!\shortuparrow$
        & HOTA$\!\shortuparrow$ & MOTA$\!\shortuparrow$ & IDF1$\!\shortuparrow$ \\
        \midrule
        &
        & 78.7 & 95.8 & 80.9
        & 57.1 & 89.3 & 58.2 \\
        $\checkmark$ &
        & 78.8 & 95.8 & 81.0
        & 57.4 & 89.4 & 58.8 \\
        $\checkmark$ & $\checkmark$
        & \bf 78.9 & \bf 95.8 & \bf 81.2
        & \bf 57.6 & \bf 89.4 &\bf  59.4 \\
        \bottomrule
    \end{tabular}}
    \label{tab:pos}
\end{table}

\begin{table}[t]
    \small
    \centering
    \caption{Ablations on the input frame length. The $T$ denotes the length of the input frame sequence.}
    \resizebox{0.8\columnwidth}{!}{
    \begin{tabular}{c
    |c@{\hspace{1cm}}c@{\hspace{1cm}}c@{\hspace{1cm}}c
    |c@{\hspace{1cm}}c@{\hspace{1cm}}c@{\hspace{1cm}}c}
        \toprule
        PuTR & \multicolumn{4}{c|}{SportsMOT val} & \multicolumn{4}{c}{DanceTrack val}  \\
        $T$
        & HOTA$\!\shortuparrow$ & MOTA$\!\shortuparrow$ & IDF1$\!\shortuparrow$ & FPS$\!\shortuparrow$
        & HOTA$\!\shortuparrow$ & MOTA$\!\shortuparrow$ & IDF1$\!\shortuparrow$ & FPS$\!\shortuparrow$ \\
        \midrule
        5
        & 76.6 & 95.7 & 77.2 & \bf 78
        & 56.2 & 89.4 & 55.3 & \bf 77 \\
        15
        & 78.0 & 95.8 & 79.6 & 77
        & 57.8 & \bf 89.6 & 59.3 & 76 \\
        30
        & 78.9 & \bf 95.8 & 81.2 & 77
        & 57.6 & 89.4 & 59.4 & 75 \\
        45
        & 79.1 & 95.7 & 81.8 & 76
        & \bf 58.3 & 89.3 & \bf 60.8 & 74 \\
        60
        & 79.1 & 95.7 & 82.0 & 75
        & 56.5 & 89.2 & 58.3 & 73 \\
        90
        & 79.0 & 95.7 & 82.2 & 73
        & 56.6 & 89.0 & 58.6 & 71 \\
        120
        &\bf  79.4 & 95.7 & \bf 83.0 & 70
        & 56.2 & 88.8 & 58.5 & 68 \\
        \bottomrule
    \end{tabular}}
    \label{tab:flength}
\end{table}

\begin{table}[t]
    \small
    \centering
    \caption{Ablations on the structure parameters. The $d_\textit{model}$ and layers denote the model dimension and the number of layers in PuTR, respectively.}
    \resizebox{0.8\columnwidth}{!}{
    \begin{tabular}{c@{\hspace{1cm}}c
    |c@{\hspace{1cm}}c@{\hspace{1cm}}c
    |c@{\hspace{1cm}}c@{\hspace{1cm}}c}
        \toprule
        \multicolumn{2}{c|}{PuTR} & \multicolumn{3}{c|}{SportsMOT val} & \multicolumn{3}{c}{DanceTrack val}  \\
        $d_\textit{model}$ & layers
        & HOTA$\!\shortuparrow$ & MOTA$\!\shortuparrow$ & IDF1$\!\shortuparrow$
        & HOTA$\!\shortuparrow$ & MOTA$\!\shortuparrow$ & IDF1$\!\shortuparrow$ \\
        \midrule
        256 & 3
        & \bf 79.0 & \bf 95.8 & \bf 81.3
        & 56.5 & 89.4 & 57.6 \\
        256 & 6
        & 78.8 & 95.8 & 81.0
        & 57.3 & 89.5 & 59.0 \\
        256 & 9
        & 78.9 & 95.8 & 81.3
        & 56.6 & 89.3 & 58.2 \\
        \midrule
        512 & 3
        & 78.9 & 95.8 & 81.2
        &\bf  58.3 & 89.4 & \bf 60.2 \\
        512 & 6
        & 78.9 & 95.8 & 81.2
        & 57.6 & 89.4 & 59.4 \\
        512 & 9
        & 78.7 & 95.8 & 81.0
        & 57.4 &\bf  89.6 & 59.0 \\
        \bottomrule
    \end{tabular}}
    \label{tab:struct}
\end{table}


\begin{figure*}[t]
    \centering
    \includegraphics[width=0.95\textwidth]{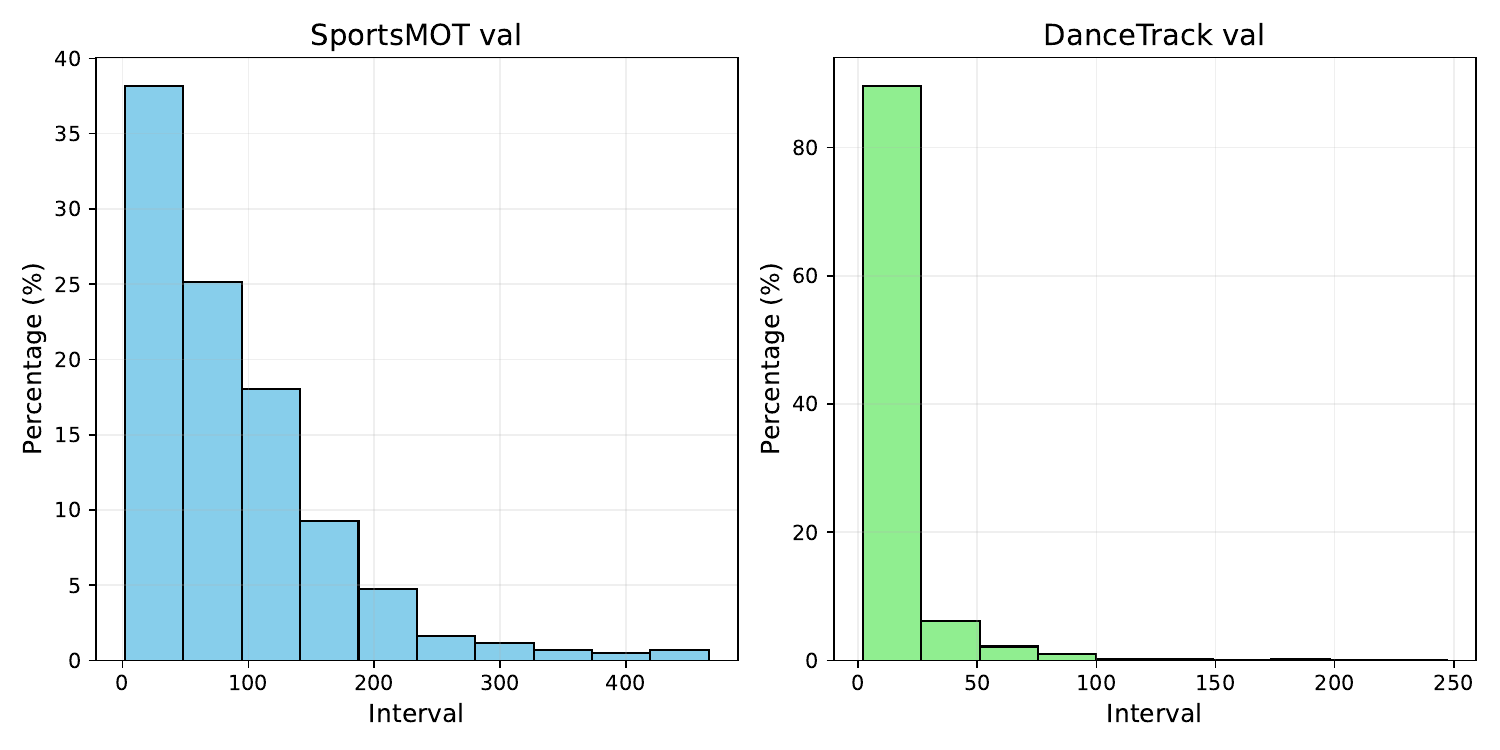}
    \caption{
        Distribution of object disappearance intervals in the SportsMOT and DanceTrack validation datasets. For each tracked identity, we compute the temporal gap (Interval, the x-axis) between adjacent detections by calculating the difference between current and previous frame numbers, while the y-axis shows their percentage among all non-consecutive cases (interval > 1).
    }
    \label{fig:interval}
\end{figure*}

\subsection{Ablation Studies}
We conduct comprehensive ablation studies on the SportsMOT and DanceTrack datasets, leveraging their extensive training data and official validation sets.

\subsubsection{Affinity Matrix}
We analyze the contribution of different components in the affinity matrix, with results presented in Tab.\ref{tab:am}.
In the first row, our baseline implementation of PuTR achieves competitive performance on both validation sets using the maximum operation alone.
Experimental results demonstrate that each additional component contributes incrementally to the overall tracking performance.
The performance gains are particularly notable on DanceTrack, where improvements exceed those observed on SportsMOT.
This differential impact can be attributed to DanceTrack's higher object density and inter-object similarity, which introduces additional complexity in the feature-based similarity matrix.
These characteristics necessitate more sophisticated compensation mechanisms to effectively filter invalid matches during bipartite matching.

\subsubsection{Attention Mask}
Tab.\ref{tab:mask} presents our comparative analysis of the proposed frame causal mask against the conventional causal mask.
We train and evaluate models using consistent mask types, incorporating random shuffling of current frame object orders during inference.
While the association results should theoretically remain invariant to object ordering in the current frame, the conventional causal mask intrinsically couples the training object order with feature extraction, introducing suboptimal constraints for the association task.
This coupling manifests in performance degradation, as evidenced by the inconsistent results under shuffled object orders in the first two rows.
In contrast, our frame causal mask maintains permutation invariance for objects within the same frame, effectively decoupling object order from association results.
The last two rows demonstrate consistent performance across different object orderings, highlighting the stability and reliability of our approach in maintaining consistency between the trajectory graph and Transformer architecture.

\subsubsection{Positional Encoding}
The effectiveness of our modified positional encodings is quantitatively demonstrated in Tab.\ref{tab:pos}.
Experimental results confirm that each encoding component contributes meaningfully to performance improvements across both datasets.
The impact is particularly pronounced on DanceTrack, where the high density and visual similarity of dancers necessitate robust temporal and spatial information to effectively distinguish between visually similar objects.
This requirement explains the more substantial improvements observed on DanceTrack compared to SportsMOT.

\subsubsection{Input Frame Length}
We investigate the impact of varying input frame lengths on tracking performance, as shown in Tab.\ref{tab:flength}.
Increasing the frame length from $5$ to $120$ frames consistently improves the association metrics (IDF1 and HOTA) on SportsMOT, reaching peak performance at $120$ frames. In contrast, DanceTrack achieves optimal performance at $45$ frames.

This performance difference stems from distinct camera setups between datasets. SportsMOT employs dynamic cameras that follow competition highlights, causing frequent object entries and exits from the frame. DanceTrack uses fixed cameras capturing the entire stage, where object disappearances mainly result from inter-object occlusions. As illustrated in Fig.\ref{fig:interval}, SportsMOT exhibits significantly longer disappearance intervals ($87$ frames on average) compared to DanceTrack ($12$ frames on average), necessitating extended temporal context for identity maintenance.

Notably, Fig.\ref{fig:interval} only depicts non-consecutive intervals, while consecutive tracking (interval = $1$) dominates modern MOT datasets, accounting for over $99\%$ of all cases. This prevalence of short-term connections explains the historical success of heuristic tracking methods. Guided by our intuitive idea, our proposed PuTR naturally unifies both short-term and long-term tracking within a single Transformer architecture while maintaining efficient inference speeds of $70$ FPS on SportsMOT and $68$ FPS on DanceTrack with $120$-frame inputs.

\subsubsection{Structure Parameters}
We analyze the influence of structure parameters on tracking performance, as shown in Tab.\ref{tab:struct}. Our experiments reveal that performance varies with the number of layers and model dimension. On SportsMOT, optimal performance is achieved with $3$ layers and a model dimension of $256$, while DanceTrack performs best with $3$ layers and a dimension of $512$.

However, rather than extensively tuning every hyperparameters for each dataset, we prioritize validating our core methodology through a straightforward implementation of PuTR.
\textit{In subsequent experiments, we maintain consistent hyperparameters across datasets as much as possible to demonstrate the solid performance of our approach, even though this will prevent PuTR from achieving optimal performance on each dataset.}

\begin{table*}
\caption{Comparisons with other foundational online methods. The \textcolor{babyblue!95}{blue background} indicates the use of the same detector, where results highlighted in \textbf{bold} represents the best results within each column, respectively. PuTR$_{120}$ denotes the model tested with a longer sequence length of $120$ frames.
The offline interpolation and the per-sequence thresholds in ByteTrack and OC-SORT are removed for fair comparison.}
\centering
\begin{minipage}{\linewidth}
    \centering
    \subfloat[SportsMOT test set]{
        \resizebox{0.47\linewidth}{!}{
        \setlength{\tabcolsep}{0.05cm}
            \begin{tabular}{
                @{\hspace{0.05cm}}l@{\hspace{0.05cm}}
                c@{\hspace{0.05cm}}
                @{\hspace{0.05cm}}c@{\hspace{0.05cm}}
                @{\hspace{0.05cm}}c@{\hspace{0.05cm}}
                @{\hspace{0.05cm}}c@{\hspace{0.05cm}}
                @{\hspace{0.05cm}}c@{\hspace{0.05cm}}
            }
            \toprule
    Method& HOTA$\!\shortuparrow$ & MOTA$\!\shortuparrow$ & IDF1$\!\shortuparrow$ & DetA$\!\shortuparrow$ & AssA$\!\shortuparrow$ \\
            \midrule
            \textit{\textbf{Detector-based}}:&&&&&\\

            Tracktor++~\cite{tracktor}&-&-&-&-&-\\

            CenterTrack~\cite{centertrack}&62.7&90.8&60.0&82.1&48.0\\

            FairMOT~\cite{fairmot}&49.3&86.4&53.5&70.2&34.7\\

            QDTrack~\cite{qdtrack}&60.4&90.1&62.3&77.5&47.2\\

            TrackFormer~\cite{trackformer}&-&-&-&-&-\\

            MOTR~\cite{motr}&-&-&-&-&-\\

            MOTIP~\cite{motip}&71.9&92.9&75.0&83.4&62.0\\
            \midrule
            \textit{\textbf{Heuristic}}:&&&&&\\
            \rowcolor{babyblue!30}DeepSORT~\cite{deepsort}&-&-&-&-&-\\
            \rowcolor{babyblue!30}ByteTrack~\cite{bytetrack}&64.1&95.9&71.4&78.5&52.3\\
            \rowcolor{babyblue!30}OC-SORT~\cite{ocsort}&73.7&96.5&74.0&88.5&61.5\\

            \aboverulesepcolor{babyblue!30}
            \midrule
            \belowrulesepcolor{babyblue!30}
            \rowcolor{babyblue!30}PuTR (\textbf{Ours})&76.0&97.1&77.1&89.3&64.8\\
            \rowcolor{babyblue!30}PuTR$_{120}$ (\textbf{Ours})&\bf 76.8&\bf 97.1&\bf 79.5&\bf 89.2&\bf 66.2\\
            \aboverulesepcolor{babyblue!30}
            \bottomrule
        \end{tabular}
        }
    }
    \hspace{0.001\textwidth}
    \subfloat[DanceTrack test set]{
        \resizebox{0.47\linewidth}{!}{
        \setlength{\tabcolsep}{0.05cm}
            \begin{tabular}{
                @{\hspace{0.05cm}}l@{\hspace{0.05cm}}
                c@{\hspace{0.05cm}}
                @{\hspace{0.05cm}}c@{\hspace{0.05cm}}
                @{\hspace{0.05cm}}c@{\hspace{0.05cm}}
                @{\hspace{0.05cm}}c@{\hspace{0.05cm}}
                @{\hspace{0.05cm}}c@{\hspace{0.05cm}}
            }
            \toprule
   Method& HOTA$\!\shortuparrow$ & MOTA$\!\shortuparrow$ & IDF1$\!\shortuparrow$ & DetA$\!\shortuparrow$ & AssA$\!\shortuparrow$ \\
            \midrule
            \textit{\textbf{Detector-based}}:&&&&&\\

            Tracktor++~\cite{tracktor}&-&-&-&-&-\\

            CenterTrack~\cite{centertrack}&41.8&86.8&35.7&78.1&22.6\\

            FairMOT~\cite{fairmot}&39.7&82.2&40.8&66.7&23.8\\

            QDTrack~\cite{qdtrack}&54.2&87.7&50.4&80.1&36.8\\

            TrackFormer~\cite{trackformer}&-&-&-&-&-\\

            MOTR~\cite{motr}&54.2&79.7&51.5&73.5&40.2\\

            MOTIP~\cite{motip}&67.5&90.3&72.2&79.4&57.6\\
            \midrule
            \textit{\textbf{Heuristic}}:&&&&&\\
            \rowcolor{babyblue!30}DeepSORT~\cite{deepsort}&45.6&87.8&47.9&71.0&29.7\\
            \rowcolor{babyblue!30}ByteTrack~\cite{bytetrack}&47.7&89.6&53.9&71.0&32.1\\
            \rowcolor{babyblue!30}OC-SORT~\cite{ocsort}&55.1&92.0&54.6&80.3&38.3\\
            \aboverulesepcolor{babyblue!30}
            \midrule
            \belowrulesepcolor{babyblue!30}
            \rowcolor{babyblue!30}PuTR (\textbf{Ours})&\bf 60.6&\bf 92.3&\bf 61.7&\bf 82.6&\bf 44.6\\
            \aboverulesepcolor{babyblue!30}
            \bottomrule
        \end{tabular}
        }
    }

\end{minipage}

\vspace{0.01cm}
\hfill

\begin{minipage}{\linewidth}
    \centering
    \subfloat[MOT17 test set]{
        \resizebox{0.47\linewidth}{!}{
        \setlength{\tabcolsep}{0.05cm}
            \begin{tabular}{
                @{\hspace{0.05cm}}l@{\hspace{0.05cm}}
                c@{\hspace{0.05cm}}
                @{\hspace{0.05cm}}c@{\hspace{0.05cm}}
                @{\hspace{0.05cm}}c@{\hspace{0.05cm}}
                @{\hspace{0.05cm}}c@{\hspace{0.05cm}}
                @{\hspace{0.05cm}}c@{\hspace{0.05cm}}
            }
            \toprule
            Method& HOTA$\!\shortuparrow$ & MOTA$\!\shortuparrow$ & IDF1$\!\shortuparrow$ & DetA$\!\shortuparrow$ & AssA$\!\shortuparrow$ \\
            \midrule
            \textit{\textbf{Detector-based}}:&&&&&\\

            Tracktor++~\cite{tracktor}&42.1&53.5&52.3&42.9&41.7\\

            CenterTrack~\cite{centertrack}&52.2&67.8&64.7&53.8&51.0\\

            FairMOT~\cite{fairmot}&59.3&73.7&72.3&60.9&58.0\\

            QDTrack~\cite{qdtrack}&53.9&68.7&66.3&55.6&52.7\\

            TrackFormer~\cite{trackformer}&57.3&74.1&68.0&60.9&54.1\\

            MOTR~\cite{motr}&57.2&71.9&68.4&58.9&55.8\\

            MOTIP~\cite{motip}&59.2&75.5&71.2&62.0& 56.9\\
            \midrule
            \textit{\textbf{Heuristic}}:&&&&&\\
            \rowcolor{babyblue!30}DeepSORT~\cite{deepsort} &61.2 &78.0& 74.5 & 63.1 & 59.7 \\
            \rowcolor{babyblue!30}ByteTrack~\cite{bytetrack}  &\bf 62.8 &\bf 78.9& \bf 77.2 & 63.8 & \bf 62.2 \\
            \rowcolor{babyblue!30}OC-SORT~\cite{ocsort} & 61.7 & 76.0 &76.2& 61.6 &62.0  \\
            \aboverulesepcolor{babyblue!30}
            \midrule
            \belowrulesepcolor{babyblue!30}
            \rowcolor{babyblue!30}PuTR (\textbf{Ours})& 62.1& 78.8 & 75.6& \bf 64.0 & 60.5\\
            \aboverulesepcolor{babyblue!30}
            \bottomrule
        \end{tabular}
        }
    }
    \hspace{0.001\textwidth}
    \subfloat[MOT20 test set]{
        \resizebox{0.47\linewidth}{!}{
        \setlength{\tabcolsep}{0.05cm}
            \begin{tabular}{
                @{\hspace{0.05cm}}l@{\hspace{0.05cm}}
                c@{\hspace{0.05cm}}
                @{\hspace{0.05cm}}c@{\hspace{0.05cm}}
                @{\hspace{0.05cm}}c@{\hspace{0.05cm}}
                @{\hspace{0.05cm}}c@{\hspace{0.05cm}}
                @{\hspace{0.05cm}}c@{\hspace{0.05cm}}
            }
            \toprule
            Method& HOTA$\!\shortuparrow$ & MOTA$\!\shortuparrow$ & IDF1$\!\shortuparrow$ & DetA$\!\shortuparrow$ & AssA$\!\shortuparrow$ \\
            \midrule
            \textit{\textbf{Detector-based}}:&&&&&\\

            Tracktor++~\cite{tracktor}&42.1&52.7&52.6&42.3&42.0\\

            CenterTrack~\cite{centertrack}&-&-&-&-&-\\

            FairMOT~\cite{fairmot} &54.6& 61.8&67.3&54.7&54.7\\

            QDTrack~\cite{qdtrack}&60.0&74.7&73.8&61.4&58.9\\

            TrackFormer~\cite{trackformer}& 54.7& 68.6&65.7&56.7&53.0\\

            MOTR~\cite{motr}&-&-&-&-&-\\

            MOTIP~\cite{motip}&-&-&-&-&-\\
            \midrule
            \textit{\textbf{Heuristic}}:&&&&&\\
            \rowcolor{babyblue!30}DeepSORT~\cite{deepsort}& 57.1& 71.8&69.6&59.0&55.5\\
            \rowcolor{babyblue!30}ByteTrack~\cite{bytetrack}& 60.4& 74.2& 74.5&62.0&59.9\\
            \rowcolor{babyblue!30}OC-SORT~\cite{ocsort}& 60.5& 73.1&74.4 &60.5&\bf 60.8\\
            \aboverulesepcolor{babyblue!30}
            \midrule
            \belowrulesepcolor{babyblue!30}
            \rowcolor{babyblue!30}PuTR (\textbf{Ours})&\bf 61.4& \bf 75.6&\bf 74.6 &\bf 62.7&60.4\\
            \aboverulesepcolor{babyblue!30}
            \bottomrule
        \end{tabular}
        }
    }
\end{minipage}
\label{tab:compare}
\vspace{-0.2cm}
\end{table*}

\subsection{Comparison with Foundational Online Methods}
Tab.\ref{tab:compare} presents results comparing PuTR with foundational online methods on four datasets and Tab.\ref{tab:domain} exhibits PuTR's domain adaptation ability in the domain shift test.


\begin{figure*}[h]
    \centering
    \includegraphics[width=0.95\textwidth]{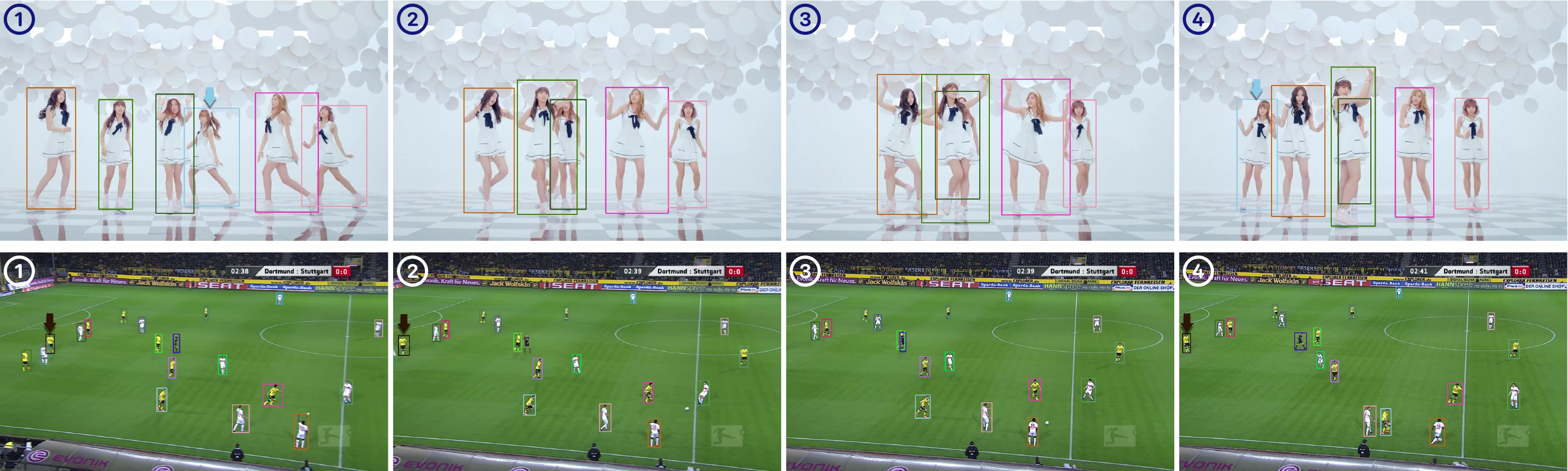}
    \caption{Visual results of PuTR on the \#dancetrack0011 video sequence of DanceTrack (top row) and the \#v\_1UDUODIBSsc\_c004 video sequence of SportsMOT (bottom row), showcasing the model's capability to handle challenging scenarios such as extended occlusions (the blue arrow), and the individual moving out of and re-entering the camera view (the dark arrow).}
    \label{fig:visual}
\end{figure*}

\subsubsection{Comparison with Heuristic Methods}
PuTR demonstrates superior performance on DanceTrack, SportsMOT\footnote{For SportsMOT, we use the official detection results following the Train+Val setup. In the Train setup, the results of PuTR are $74.3\%/95.2\%/75.7\%/87.2\%/63.4\%$ on HOTA/MOTA/IDF1/DetA/AssA.}, and MOT20, while maintaining competitive results on MOT17. Unlike heuristic approaches, PuTR employs a streamlined Transformer architecture that extracts object features and handles various associations uniformly, eliminating the need for handcrafted rules to specify different scenarios.

As illustrated in Fig.\ref{fig:visual}, PuTR effectively models long-range dependencies through its Transformer architecture. This capability enables robust tracking even during extended occlusions and significant position changes, without relying on any motion cues. For instance, the dancer with blue arrow in the top row is successfully tracked despite prolonged occlusion and substantial spatial displacement. Similarly, PuTR maintains tracking continuity when subjects exit and re-enter the camera view, as demonstrated by the athlete with dark arrow in the bottom row.

This approach inherently addresses challenges that typically require explicit handling in heuristic methods, such as non-linear motion~\cite{fasttrack}, low frame rates~\cite{coltrack}, and camera motion~\cite{ucmctrack}. On DanceTrack, despite complex choreographic interactions, PuTR surpasses the second-best method, OC-SORT, with an IDF1 score of $61.7\%$ versus $54.6\%$. When extending the consecutive frame window to $T=120$ on SportsMOT, PuTR$_{120}$ achieves remarkable scores of $76.8\%/97.1\%/79.5\%$ on  HOTA/MOTA/IDF1, highlighting its robust long-term tracking capabilities.

\begin{figure*}[t]
    \centering
    \includegraphics[width=0.95\textwidth]{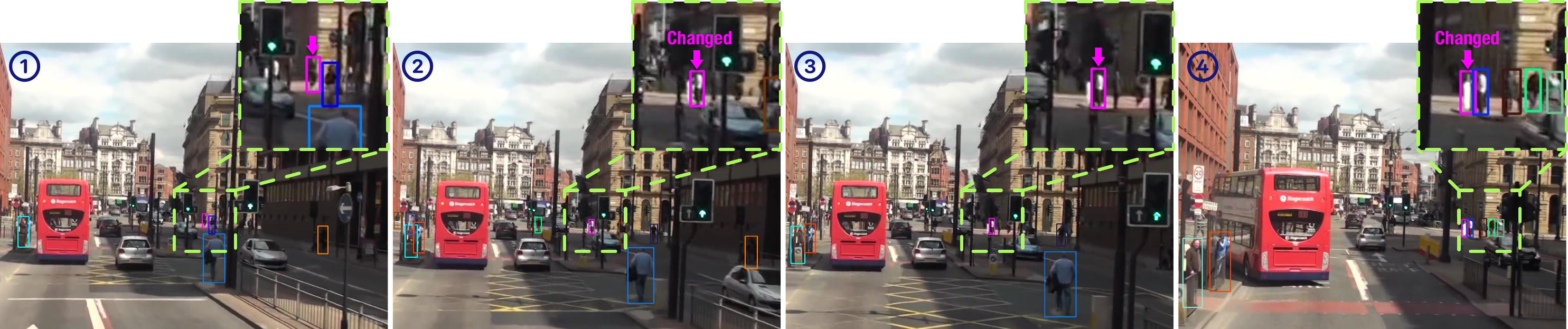}
    \caption{An example of the failure case in video sequence \#MOT17-14 of MOT17. The tiny pink bbox changes individuals twice in the second and fourth frame, due to the weak appearance cues.}
    \label{fig:failure}
\end{figure*}

However, PuTR shows limitations in processing small-scale objects especially in the low-resolution images of MOT17, as depicted in Fig.\ref{fig:failure}. The abundance of tiny bboxes in MOT17 poses challenges for feature-based association, while motion models handle such cases more effectively. This limitation explains why PuTR, like other detector-based approaches, falls short of heuristic methods' performance on MOT17.

\subsubsection{Comparison with Detector-based Methods} PuTR outperforms the representative JDE~\cite{fairmot, qdtrack}, CNN-based~\cite{tracktor, centertrack}, and DETR-based~\cite{trackformer, motr} methods on all benchmarks, except the latest DETR-based approach MOTIP~\cite{motip} warranting a detailed analysis.

MOTIP leverages its methodology's advantages but requires additional training data and substantial computational resources (typically $8$ GPUs for days) to achieve satisfactory performance, hence excelling on DanceTrack and SportsMOT. For MOT17, the CrowdHuman~\cite{crowdhuman} dataset, containing massive human images, is incorporated following prior work~\cite{motr, memotr} to mitigate overfitting, improving performance while increasing training complexity. For MOT20, where the average pedestrian count per frame exceeds DETR's detection capacity, only the early TrackFormer~\cite{trackformer} is evaluated by modifying Deformable-DETR, while later incremental works focus on datasets with more moderate object query numbers.
In contrast, PuTR performs streamlined training from scratch on a single GPU in $1$ hour\footnote{Specifically, training time on DanceTrack, SportsMOT, MOT17, and MOT20 is $60$, $40$, $30$, and $35$ minutes.} without any additional data or pretraining, regardless of dataset scale or object count.

PuTR demonstrates superior performance on SportsMOT compared to MOTIP, while exhibiting potential for improvement on DanceTrack. The performance gap on DanceTrack can be attributed to the inherent complexity of dancer interactions, where the YOLOX detector encounters challenges in generating accurate detections under severe occlusion and crowded scenarios.
Analysis of the HOTA metric reveals PuTR's superior performance compared to other heuristic approaches, indicating robust handling of detection inaccuracies. However, the lower AssA score relative to MOTIP suggests that MOTIP's end-to-end approach with shared ID dictionary is more effective in maintaining identity consistency and filtering inaccurate detections in extreme interaction scenarios. Nevertheless, MOTIP's advantages come at the cost of increased training complexity and reduced domain adaptability.
These observations underscore a fundamental trade-off in MOT: no single methodology can optimally address all MOT challenges simultaneously. By leveraging precise detections, PuTR maximizes its association capabilities, achieving state-of-the-art results on SportsMOT.  Notably, even with additional image data, MOTIP fails to surpass PuTR's performance on MOT17. Furthermore, PuTR demonstrates strong scalability, extending effectively to dense scenarios in MOT20.

In the inference efficiency, PuTR demonstrates remarkable inference speed. Unlike MOTIP's $22$ FPS\footnote{We test the official model's speed on the same machine as PuTR and exceed the original $16$ FPS report.} on DanceTrack, SportsMOT, and MOT17, PuTR can process video sequences at a much higher frame rate due to its relatively simple architecture. Specifically, PuTR's average FPS on the sparse DanceTrack, SportsMOT, and MOT17 datasets are $75$, $77$, and $62$, respectively. Even on the dense MOT20 dataset, it can still achieve an impressive $25$ FPS. While the massive number of tracked objects per frame in MOT20 can lengthen the input object sequence and consequently slow down the speed, it still meets real-time requirements. Moreover, the separated approach offers the advantage of independent execution for detection and association, enabling parallel processing pipelines for further efficiency enhancement.


\begin{table*}
    \small
    \centering
    \caption{
    Domain shift analysis of PuTR across different datasets. Models trained on individual datasets (indicated by subscripts) are evaluated on all test sets. $\Delta_{\textit{max}}$ denotes the maximum performance gap observed across the four test sets, reflecting the model's domain adaptation capability.
    }
    \resizebox{0.95\textwidth}{!}{
    \renewcommand{\arraystretch}{1.2}
    \begin{tabular}{l
    |c@{\hspace{0.2cm}}c@{\hspace{0.2cm}}c
    |c@{\hspace{0.2cm}}c@{\hspace{0.2cm}}c
    |c@{\hspace{0.2cm}}c@{\hspace{0.2cm}}c
    |c@{\hspace{0.2cm}}c@{\hspace{0.2cm}}c}
        \toprule

        \multirow{2}{*}{Method}
        & \multicolumn{3}{c|}{SportsMOT test}
        & \multicolumn{3}{c|}{DanceTrack test}
        & \multicolumn{3}{c|}{MOT17 test}
        & \multicolumn{3}{c}{MOT20 test}  \\
        & HOTA$\!\shortuparrow$ & MOTA$\!\shortuparrow$ & IDF1$\!\shortuparrow$
        & HOTA$\!\shortuparrow$ & MOTA$\!\shortuparrow$ & IDF1$\!\shortuparrow$
        & HOTA$\!\shortuparrow$ & MOTA$\!\shortuparrow$ & IDF1$\!\shortuparrow$
        & HOTA$\!\shortuparrow$ & MOTA$\!\shortuparrow$ & IDF1$\!\shortuparrow$ \\
        \midrule
        PuTR$_{\textit{MOT17}}$
        & 75.7 & 97.0 & 76.4
        & 59.5 & 92.2 & 60.2
        & 62.1 & 78.8 & 75.6
        & 61.3 & 75.6 & 74.5 \\
        PuTR$_{\textit{MOT20}}$
        & 75.9 & 97.1 & 77.0
        & 58.9 & 92.2 & 59.4
        & 62.1 & 78.8 & 75.6
        & 61.4 & 75.6 & 74.6 \\
        PuTR$_{\textit{DanceTrack}}$
        & 75.9 & 97.1 & 76.6
        & 60.6 & 92.3 & 61.7
        & 61.9 & 78.9 & 75.3
        & 61.1 & 75.4 & 74.2 \\
        PuTR$_{\textit{SportsMOT}}$
        & 76.0 & 97.1 & 77.1
        & 59.4 & 92.2 & 60.2
        & 61.9 & 79.0 & 75.5
        & 61.4 & 75.6 & 74.8 \\
        \midrule
        $\Delta_{\textit{max}}$
        & 0.3 & 0.1 & 0.7
        & 1.7 & 0.1 & 2.3
        & 0.2 & 0.2 & 0.3
        & 0.3 & 0.2 & 0.6 \\
        \bottomrule
    \end{tabular}}
    \label{tab:domain}
\end{table*}
\subsubsection{Domain Adaptation Ability} Robust association ability across various scenarios is crucial for MOT methods. Heuristic methods can tune hyperparameters to adapt to most common cases, while other learning-based methods rarely report their domain adaptation ability across datasets despite the urge to address this issue. Only the work~\cite{darth} specifically addresses the domain adaptation problem in MOT, reporting that QDTrack exhibits a striking degradation, decreasing by $49\%$ in IDF1 and $35\%$ in HOTA when evaluated on DanceTrack with weights trained on MOT17. We also evaluate the official MOTIP weights trained on MOT17 on the DanceTrack test set, only yielding $44.9\%$ IDF1 and $43.0\%$ HOTA.

In contrast, from the domain shift analysis in Tab.\ref{tab:domain}, we find that PuTR has excellent domain adaptation ability across domains. Without any other operation, just applying the weights trained on one dataset to another, the maximum cross-dataset gap is only $2.3\%$ in IDF1 and $1.7\%$ in HOTA, which has never been reported in previous learning-based works.
This property of PuTR can be attributed to the fact that it takes advantage of different methodologies, which is appropriate in terms of methodological cross-pollination and integration.
It is the separated manner that PuTR can obtain precise detection results like heuristic methods, and the self-attention mechanism of the Transformer that can discriminate individuals in the feature space, even in scenes that have never been trained. Furthermore, the affinity matrix also plays a crucial role, as it circumvents concerns regarding the dictionary capacity or the exhaustion issue in MOTIP. This property makes the new direction we proposed more fascinating, and we believe it will attract more attention in the future and promote a new development in the MOT field.

\section{Discussion}
Through above extensive experiments and analyzes, we have demonstrated PuTR's effectiveness across diverse tracking scenarios. While there remains a performance gap between PuTR and state-of-the-art incremental methods, our primary contribution lies in establishing the viability of a pure Transformer architecture for modeling the association problem in MOT. The experimental results validate this architectural direction and reveal the potential of Transformer-based approaches for tackling MOT association challenges. Besides the ground-based scenarios, PuTR may also be applicable to UAV tracking tasks~\cite{he2024temporal, xu2025online, yao2023folt}, as the capability to track objects from fast-moving cameras shown in the SportsMOT dataset suggests.

Furthermore, we outline potential future directions for exploration:
\begin{itemize}
\item \textbf{Model Capacity.}
The results in Tab.\ref{tab:struct} show that simply increasing the model size does not consistently improve PuTR's performance. Despite using the same decoder-only architecture as Large Language Models (LLMs), effective scaling of PuTR requires further investigation to achieve the impressive perceptual capabilities of modern LLMs.

\item \textbf{Motion Cues.} Our current method solely employs the coordinate information of objects. However, as illustrated in Fig.\ref{fig:failure}, PuTR struggles with false matches between small objects in the MOT17 dataset due to insufficient appearance cues for re-identification. Incorporating motion cues, such as displacement and velocity, could alleviate this issue by providing additional discriminative information, especially in challenging scenarios involving tiny objects.
\end{itemize}

In summary, this work pioneers an innovative Transformer-based direction for MOT. The self-attention-driven modeling of trajectories showcased in PuTR opens up exciting possibilities for further advancements in this crucial computer vision task.

\section{Conclusion}
In this work, we introduce a novel idea to MOT by demonstrating that trajectory graphs can be effectively modeled using a classical Transformer architecture in a separated and online manner. This idea successfully unifies short- and long-term tracking within a single framework. We validate this concept through our proposed Pure Transformer (PuTR) architecture, which achieves competitive performance while maintaining computational efficiency. Extensive experimental results across multiple datasets demonstrate that PuTR not only achieves solid tracking performance but also exhibits robust domain adaptation capabilities. Our findings establish a promising foundation for future MOT research.

\medskip

\small
\bibliographystyle{plain}
\bibliography{main}


\appendix
\newpage
\appendix
\begin{center}
\bf{\huge Appendix}
\\
\end{center}

\section{Inference Pipeline}

\label{app:pipe}

\begin{algorithm}
\caption{Inference pipeline.}
\label{algo:pipe}
\begin{algorithmic}[1]
\Require A video sequence $\mathcal{V}$; a detector $\textit{Det}(\cdot)$; a lost tolerance $T$.

\State The set of trajectories $\mathcal{T^*} \Leftarrow \emptyset$; The set of active trajectories $\mathcal{T}_\textit{active}^* \Leftarrow \emptyset$. \;
\For{$\textit{frame}_t$ {\rm \bf in} $\mathcal{V}$}
    \Statex\ /* generate detected objects */
    \State $\mathcal{O}_t \Leftarrow \textit{Det}(\textit{frame}_t)$; \;
    \Statex\ /* generate an affinity matrix */
    \State $\mathbf{A}\cdot\mathbf{W} \Leftarrow \textit{GenAffinity}(\mathcal{T_\textit{active}^*}, \mathcal{O}_t)$; \;
    \Statex\ /* associate */
    \State $\mathcal{O}_{m}, \mathcal{T}_{m}^*, \mathcal{O}_{u}, \mathcal{T}_{u}^* \Leftarrow \textit{Hungarian}(\mathbf{A}\cdot\mathbf{W})$
    \State $\mathcal{O}_{m}\! : \text{matched detections from } \mathcal{O}_{t}$; \;
    \State $\mathcal{T}_{m}^* : \text{matched tracklets from } \mathcal{T_\textit{active}^*}$; \;
    \State $\mathcal{O}_{u} : \text{unmatched detections from } \mathcal{O}_{t}$; \;
    \State $\mathcal{T}_{u}^* : \text{unmatched tracklets from } \mathcal{T_\textit{active}^*}$; \;
    \Statex\ /* update tracklets */
    \For{$\mathcal{T}_\textit{tid}, o_\textit{tid}^t \text{ \rm \bf in  } \mathcal{T}_{m}^*, \mathcal{O}_{m}$}
        \State $\mathcal{T}_\textit{tid} \Leftarrow  \textit{Update}(\mathcal{T}_\textit{tid}, o_\textit{tid}^t)$; \;
    \EndFor
    \Statex\ /* remove tracklets */
    \For{$\mathcal{T}_\textit{tid} \text{ \rm \bf in  } \mathcal{T}_{u}^*$}
        \If{$\textit{LostGap}(\mathcal{T}_\textit{tid}) > T \text{ \rm \bf or } \mathcal{T}_\textit{tid}.\textsf{State} == \textit{New}$}
            \State $\mathcal{T}_{u}^* \Leftarrow  \mathcal{T}_{u}^* \setminus \{ \mathcal{T}_\textit{tid}$ \}; \;
            \State $\mathcal{T^*} \Leftarrow  \mathcal{T}^* \cup \{ \mathcal{T}_\textit{tid} \} $; \;
        \Else
            \State $\mathcal{T}_\textit{tid}.\textsf{State} \Leftarrow \textit{Lost}$; \;
        \EndIf
    \EndFor
    \State $\mathcal{T}_\textit{active}^* \Leftarrow  \mathcal{T}_{m}^* \cup \mathcal{T}_{u}^*$; \;
    \Statex\ /* initialize new tracklets */
    \State $\mathcal{T}_\textit{active}^* \Leftarrow  \mathcal{T}_\textit{active}^* \cup$ \textit{Init}( $\mathcal{O}_{u}$); \;
\EndFor
\State $\mathcal{T}^* \Leftarrow  \mathcal{T}^* \cup \mathcal{T}_\textit{active}^* $; \;
\Ensure $\mathcal{T}^*$.
\end{algorithmic}
\end{algorithm}

Algo.\ref{algo:pipe} shows the inference pipeline of our method. The pipeline follows a simple style to SORT, requiring only one association during one frame to highlight the Transformer's capability in solving the association problem.

The inputs of the pipeline are a video sequence $\mathcal{V}$, a detector $\textit{Det}(\cdot)$, and a lost tolerance $T$ (same as the input frame length of PuTR).
The output is the set of trajectories $\mathcal{T}^*$, and we also maintain a set of active trajectories $\mathcal{T}_{\textit{active}}^*$ to record the tracklets that are not terminated yet.
Each tracklet in $\mathcal{T}_{\textit{active}}^*$ has a state attribute, called \textsf{State} (one of \textit{Track}, \textit{Lost}, or \textit{New}), to indicate its status at the current frame.

For each frame $\textit{frame}_t$ in $\mathcal{V}$, we obtain the set of detected objects $\mathcal{O}_t$ via $\textit{Det}(\cdot)$ (line 3), filtering by detection threshold $\tau_{\textit{det}}$ and initializing tracking identities $\textit{tid}$ to $-1$. We then generate an affinity matrix using $\mathcal{T}_{\textit{active}}^*$ and $\mathcal{O}_t$ through PuTR as described in the main paper (line 4).
The Hungarian algorithm then processes $\mathbf{A}\cdot\mathbf{W}$ to obtain the assignment results, i.e., matched objects $\mathcal{O}_{m}$, matched tracklets $\mathcal{T}_{m}^*$, unmatched detections $\mathcal{O}_{u}$, and unmatched tracklets $\mathcal{T}_{u}^*$ (lines 5 to 9).
The objects in $\mathcal{O}_{m}$ will be updated into the corresponding tracklets in $\mathcal{T}_{m}^*$ (lines 10 to 12), where the \textsf{State} of tracklets will be set to \textit{Track}.
Then, the tracklets in $\mathcal{T}_{u}^*$ will be moved from $\mathcal{T}_{\textit{active}}^*$ into $\mathcal{T}^*$ if the lost gap is larger than $T$ or its \textsf{State} is \textit{New}, following ByteTrack. Otherwise, they will remain in $\mathcal{T}_{\textit{active}}^*$, and their \textsf{State} will be set to \textit{Lost} (lines 13 to 20).
Next, objects in $\mathcal{O}_{u}$ will be initialized as new tracklets and added into $\mathcal{T}_{\textit{active}}^*$ after filtering with the tracklet threshold $\tau_{\textit{new}}$ (line 22).
Finally, after processing all frames in $\mathcal{V}$, $\mathcal{T}_\textit{active}^*$ will be merged into $\mathcal{T}^*$, and $\mathcal{T}^*$ is the final tracking result of the video sequence $\mathcal{V}$ (line 24).


\section{Similarity Compensation}
\label{app:similarityc}
We elaborate on the IoU matrices $\mathbf{I}^{\alpha}$ and $\mathbf{I}^{\beta}$ used for similarity compensation in the similarity matrix $\mathbf{S}$.

The design of $\mathbf{I}^{\alpha}$ and $\mathbf{I}^{\beta}$ for similarity matrix compensation is based on a key assumption in heuristic methods: objects belonging to the same individual should exhibit high similarity in both appearance and motion cues across adjacent frames. This implies that for the same individual, high similarity must be present simultaneously in both feature space and bbox scale space, while bbox positions may vary due to rapid movement.

For scale compensation, we utilize only the bbox scale in $\mathbf{I}^{\alpha}$ to compensate for the feature similarity in the scale space:
\begin{equation}
\mathbf{I}^{\alpha}_{ij} = \textit{IoU'}(\textit{bbox}_i, \textit{bbox}_j),
\end{equation}
where $\textit{bbox}_i$ and $\textit{bbox}_j$ are the bboxes of the $i$-th object in $[\mathcal{O}_{t-T}, ..., \mathcal{O}_{t-1}]$ and the $j$-th object in $\mathcal{O}_t$, respectively. A bbox is defined as $[u, v, w, h]$, where $u$ and $v$ are the center coordinates, and $w$ and $h$ are the width and height. $IoU'(\cdot)$ denotes the IoU score calculated after setting the center coordinates of both bboxes to zero while preserving their width and height. This alignment ensures that the IoU score is not affected by the bbox positions, focusing solely on the object scale. By multiplying $\mathbf{I}^{\alpha}$ in $\mathbf{S}$, we emphasize the feature similarity of objects with similar scales.

For position compensation, we leverage the fact that objects appear in close proximity in consecutive frames. We compute the standard IoU score ($\mathbf{I}^{\beta}$) using detections and the last positions of trajectories, following heuristic methods but without introducing any motion model. When computing $\mathbf{I}^{\beta}_{ij}$, $\textit{bbox}_i$ represents the bbox of the trajectory of the $i$-th object in $[\mathcal{O}_{t-T}, ..., \mathcal{O}_{t-1}]$, while $\textit{bbox}_j$ remains the bbox of the $j$-th object in $\mathcal{O}_t$. The trajectory bbox is taken from the object with the largest frame index in the same \textit{tid}, denoting the last position of the trajectory.
Since IoU evaluates to zero for disjoint bboxes — a common occurrence for fast-moving objects — we opt to add rather than multiply $\mathbf{I}^{\beta}$ in $\mathbf{S}$. This choice prevents the feature similarity from being completely suppressed by zero IoU scores, ensuring that the model can still identify objects with similar features but different positions. This approach effectively compensates for the feature similarity in the position space.

Through these compensation strategies, we aim to enhance the feature similarity in both scale and position spaces, enabling the model to better distinguish between objects with similar features but different scales or positions.

\end{document}